# Off-Line Handwritten Signature Identification Using Rotated Complex Wavelet Filters

M.S. Shirdhonkar[1] and Manesh Kokare[2]

[1] Department of Computer Science and Engineering, BLDEA's CET, Bijapur, Karnataka, India

[2] 2Department of Electronics and Telecommunication, SGGS IOT, Nanded, Maharastra, India

**Abstract**

In this paper, a new method for handwritten signature identification based on rotated complex wavelet filters is proposed. We have proposed to use the rotated complex wavelet filters (RCWF) and dual tree complex wavelet transform(DT-CWT) together to derive signature feature extraction, which captures information in twelve different directions. In identification phase, Canberra distance measure is used. The proposed method is compared with discrete wavelet transform (DWT). From experimental results it is found that signature identification rate of proposed method is superior over DWT

## 1. Introduction

1.1 Motivation

Authentication and affirmation of statements, documents, scripts etc, from times immemorial has been done through signatures. Even today, where every thing has gone digital, signature plays a vital role. They appear on many types of documents such as bank cheques, credit cheques, governmental documents, wills over assets of a person and many other documents of greater importance. But this type of authentication is also subject to mal practices and crimes. Forgery and imitation of signatures of other person may help anyone to gain access to his/her valuable assets or can lead to undesirable consequences. Identification of signatures by human eye and study may be error prone and manipulative, thus an automated document processing system that can analyze and identify a signature serves as an effective, useful and less error prone and non manipulative tool.

Signature's validity confirmation for different documents is an important problem domain in automatic document processing. An area where signature identification finds application is in banking, user login in computers or PDA (Personal digital assistant), for access control, to check for authentication of official documents etc. There are two modes for signature identification and verification: Static or off-line and Dynamic or on-line. In static mode, the input of system is a 2D image of signature. Contrary to this, in dynamic mode, the input is signature trace in time domain. In the Dynamic mode, the person puts his signature on an electronic tablet through an electronic pen. His/her obtained signature is sampled with each sample having three attributes: The 2-dimensional co-ordinates; x and y and time of sample occurrence, t. The time attribute of each sample is used to extract useful information such as start and stop points, velocity and acceleration of the signature stroke. Some electronic tablets in addition to time sampling can digitize the pressure. Such additional information in the dynamic mode increases identification rate as compared to the static mode. But Dynamic mode has a greater disadvantage: it is on-line, hence it requires presence of the person whose signature is needed and that too has been taken digitally. Thus it cannot be applied to other important cases where there is absence of person whose sign is needed and cases where analysis and identification needs to be carried out of existing documents with signature marks. Thus off-line signature verification becomes an inevitable choice and finds universal application.

1.2 Related works

Signature verification contain two areas: off-line signature verification, where signature samples are scanned into image representation and on-line signature verification, where signature samples are collected from a digitizing tablet which is capable of pen movements during the writing .In 2009, Ghandali and Moghaddam have proposed an off-line Persians signature identification and verification based on Image registration, DWT (Discrete Wavelet Transform) and fusion. They used DWT for features extraction and Euclidean distance for comparing features. It is language dependent method [1].In 2008, Larkins and Mayo have introduced a person dependent off-line signature verification method that is based on Adaptive Feature Threshold (AFT) [2].AFT enhances the method of converting a simple feature of signature to binary feature vector to improve its representative similarity with training signatures. They have used combination of spatial pyramid and equimass sampling grids to improve representation of a signature based on gradient direction. In classification phase, they used DWT and graph matching methods. In another work, Ramachandra et al [3], have proposed cross-validation for graph matching based off-line signature verification (CSMOSV) algorithm in which graph matching compares signatures and the Euclidean distance measures the dissimilarity between signatures.

In 2007, Kovari et.al [4] presented an approach for off-line signature verification, which was able to preserve and take usage of semantic information. They, used position and direction of endpoints in features extraction phase. Porwik [5] introduced a three stages method for offline signature recognition. In this approach the Hough transform ,center of gravity and horizontal-vertical signature histogram have been employed, using both static and dynamic features that were processed by DWT has been addressed in[6].The verification phase of this method is based on fuzzy net using the enhanced version of the MDF(Modified Direction feature)extractor has been presented by Armand et.al [7].The different neural classifier such as Resilient Back Propagation(RBP) neural network and Radial Basis Function(RBF)network have been used in verification phase of this method. In 2005, Chen and Srihari [8] described an approach that obtains an exterior contour of the image to define pseudo writing path. To match two signatures a dynamic time wrapping (DTW) method has been employed to segment signature into curves.

The main contribution of this paper is that, we have proposed an off-line handwritten signature identification using rotated complex wavelet filters and dual tree complex wavelet transform, which captures information in twelve different directions for identification. In identification phases Canberra distance measure is used. The experimental results of proposed method were satisfactory and found that it gives better results as compared with earlier approach. The rest of paper is organized as follows. In section 2, discusses the feature extraction phase. The signature identification approaches is presented in section 3. In section 4, the experimental results and the selection of training samples are presented, and finally section 5 concludes the work.

## 2. Feature Extraction Phase

The major task of feature extraction is to reduce image data to much smaller in size which represents the important characteristic of the image. In signature identification, edge information is very important in characterizing signature properties. Therefore we proposed the use of DT-CWT and DT-RCWF jointly, which captures the information in twelve different directions. The performance of the system is compared with standard discrete wavelet transform which captures information in only three directions.

### 2.1 Discrete Wavelet Transform Features

The multi resolution wavelet transform decomposes a signal into low pass and high pass information. The low pass information represents a smoothed version and the main body of the original data. The high pass information represents data of sharper variations and details. Discrete Wavelet Transform decomposes the image into four sub-images when one level of decomposing is used. One of these sub-images is a smoothed version of the original image corresponding to the low pass information and the other three ones are high pass information that represents the horizontal, vertical and diagonal edges of the image respectively. When two images are similar, their difference would be existed in high-frequency information. A DWT with N decomposition levels has 3N+1 frequency bands with 3N high-frequency bands [9]. The impulse response associated with 2-D discrete wavelet transform are illustrated in Fig. 1 as gray-scale image.

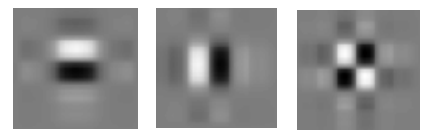

Fig.1. Impulse response of $0^0$, $90^0$ and $\pm 45^0$ of DWT

### 2.2 Dual Tree Rotated Complex Wavelet Filters

Drawbacks of the DWT are overcome by the complex wavelet transform (CWT).By introducing limited redundancy into the transform. But still it suffer from problem like no perfect reconstruction is possible using CWT decomposition beyond level 1, when input to each level becomes complex. To overcome this, Kingsbury [11] proposed a new transform, which provides perfect reconstruction along with providing the other advantages of complex wavelet, which is DT-CWT. The DT-CWT uses a dual tree of real part of wavelet transform instead using complex coefficients. This introduces a limited amount of redundancy and provides perfect reconstruction along with providing the other advantages of complex wavelets. The DT-CWT is implemented using separable transforms and by combining subband signals appropriately. Even though it is non-separable yet it inherits the computational efficiency of separable transforms. Specifically, the 1-D DT-CWT is implemented using two filter banks in parallel, operating on the same data. For d-dimensional input, a $L$ scale DT-CWT outputs an array of real scaling coefficients corresponding to the low pass subbands in each dimension. The total redundancy of the transform is $2^d$ and independent of $L$. The mechanism of the DT-CWT is not covered here. See [10], [12-13] for a comprehensive explanation of the transform and details of filter design for the trees. A complex valued $\psi(t)$ can be obtained as

$$\psi(x) = \psi_h(x) + j\psi_g(x) \qquad (1)$$

Where $\psi_h(x)$ and $\psi_g(x)$ are both real-valued wavelets. The impulse responses of six wavelets associated with 2-D dual tree complex wavelet transform are illustrated in Fig. 2.

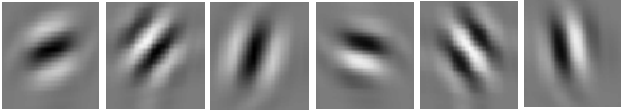

Fig. 2 Impulse response of six wavelet filters $+15^0$, $+45^0$, $+75^0$ $-15^0$, $-45^0$ and $-75^0$ of complex wavelet.

### 2.3 Dual Tree Rotated Complex Wavelet Filters

Directional 2D RCWF are obtained by rotating the directional 2D DT-CWT filters by $45^0$ so that decomposition is performed along new direction, which are apart from decomposition $45^0$ directions of CWT[10]. The size of a filter is $(2N-1)X(2N-1)$, where $N$ is the length of the 1-D filter. The decomposition of input image with 2-D RCWF followed by 2-D down sampling operation is performed up to the desired level. The computational complexity associated with RCWF decomposition is the same as that of standard 2-D CWT, if both are implemented in the 2-D frequency domain. The set of RCWFs retains the orthogonal property. The six sub bands of 2D DT-RCWF gives information strongly oriented at ($30^°, 0^°, -30^°, 60^°, 90^°, 120^°$).The mechanism of the DT-RCWF is not covered here. See [10],[12-13] for a comprehensive explanation of the transform and details of filter design for the trees. Thus, the 2D DT-CWT and RCWF provide us with more directional selectivity in the direction

$$\left\{ \begin{array}{l} (+15°, +45°, +75°, -15°, -45°, -75°), \\ (0°, +30°, +60°, +90°, 120°, -30°) \end{array} \right\} \text{ than}$$

the DWT whose directional sensitivity is in only three directions $\{0°, \pm 45°, 90°\}$. The six wavelets associated with rotated complex wavelet transform are illustrated in Fig.3.

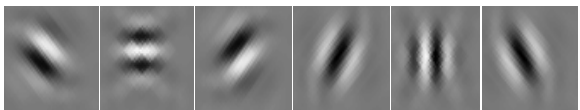

Fig. 3 Impulse response of $-30^0$, $0^0$, $+30^0$, $+60^0$, $90^0$ and $120^0$ of rotated complex wavelet

### 2.4 Feature Database Creation

To conduct the experiments, we were computed two different feature sets using *algorithm* 1 and *algorithm* 2, which uses DWT and combined DT-CWT and DT-RCWF respectively. To construct the feature vectors of each signature in the database, we decomposed each signature using DT-CWT and DT-RCWF up to 6$^{th}$ level. The Energy and Standard Deviation (STD) were computed separately on each sub band and the feature vector was formed using these two parameter values. The Energy $E_k$ and Standard Deviation $\sigma_k$ of k$^{th}$ sub band is computed as follows

$$E_k = \frac{1}{M \times N} \sum_{i=1}^{M} \sum_{j=1}^{N} |W_k(i,j)| \qquad (2)$$

$$\sigma_k = \left[ \frac{1}{M \times N} \sum_{i=1}^{N} \sum_{j=1}^{M} (W_k(i,j) - \mu_k)^2 \right]^{\frac{1}{2}} \qquad (3)$$

Where $W_k(i,j)$ is the $k^{th}$ wavelet-decomposed sub band, $MxN$ is the size of wavelet decomposed sub band, and $\mu_k$ is the mean of the $k^{th}$ sub band. The resulting feature vector using energy and standard deviation are $\bar{f}_E = [E_1 \ E_2 \ ... \ E_n]$ and $\bar{f}_\sigma = [\sigma_1 \ \sigma_2 \ ... \ \sigma_n]$ respectively. So combined feature vector is

$$\bar{f}_{\sigma\mu} = [\sigma_1 \ \sigma_2 \ ... \ \sigma_n \ E_1 \ E_2 \ ... \ E_n] \qquad (4)$$

The step by step procedure for feature database creation using discrete wavelet transform and combined DT-CWT and DT-RCWF are explained in algorithm 1 and algorithm 2 respectively.

**Algorithm 1:** Feature database creation using DWT

**Input:**
Signature image Database: DB
1D filters                : LF, HF
Handwritten Signature : $S_i$
**Output:** Feature database FV
**Begin**
    **For** *each $S_i$ in DB do*
        *Decompose the $S_i$ by applying low pass LF and high pass HF filters up to 6$^{th}$ level*
        *Calculate energy E and standard deviation SD for each subband using (2) and (3) respectively in each level*
        *Feature vector f= [E U SD]*
        *FV=FV U f*
    **End for**
**End**

---

**Algorithm 2:** Feature database creation using DT-CWT and DT-RCWF

---

**Input:**
  Signature image Database: DB
  2D DT-CWT filters         : F
  Handwritten Signature   : $S_i$
**Output:**
  Feature database            : FV
**Begin**
  **If** DT-RCWF
  Rotate 2D filters F by $45^0$
  **End if**
  **For** each $S_i$ in DB do
  Decompose the $S_i$ by applying 2D filters F up to $6^{th}$ Level. Calculate energy E and standard deviation SD for each subband using (2) and (3) respectively in each level
   Feature vector f= [E U SD]
   FV=FV U f
  **End for**
**End**

---

## 3. Signature Identification Phase

There are several ways to work out the distance between two points in multidimensional space. We have used Canberra distance metric as distance measure. If $x$ and $y$ are the feature vectors of the database and query signature, respectively, and have dimension d, then the Canberra distance is given by

$$\text{Canb}(x, y) = \sum_{i=1}^{d} \frac{|x_i - y_i|}{|x_i| + |y_i|} \qquad (5)$$

The step by step procedure of identification is as follows,

---

**Algorithm 3**: Handwritten Signature Identification

---

**Input**: Test signature: St
     Feature database: FV
**Output**: Distance vector: Dist
      Handwritten signature identification
**Begin**
    Calculate feature vector of test signature St using algorithm 1
  **For** each fv in FV do
    Dist= Calculate distance between test signature and fv using (5)
  **End for**
    Display the minimum distance signature from distance vector.
**End**

---

## 4. Experimental Results

### 4.1. Image Database

The signatures were collected using either black or blue ink (No pen brands were taken into consideration), on a white A4 sheet of paper, with eight signature per page. Signatures were scanned subsequently to digitize individual with a resolution in 256 grey levels. Images were obtained in rectangular areas of size 256x256 pixels. Sample signature image database is shown in Fig.3. A group of 52 persons are selected for 16 specimen signatures which make the total of 52x16=832 signature database.

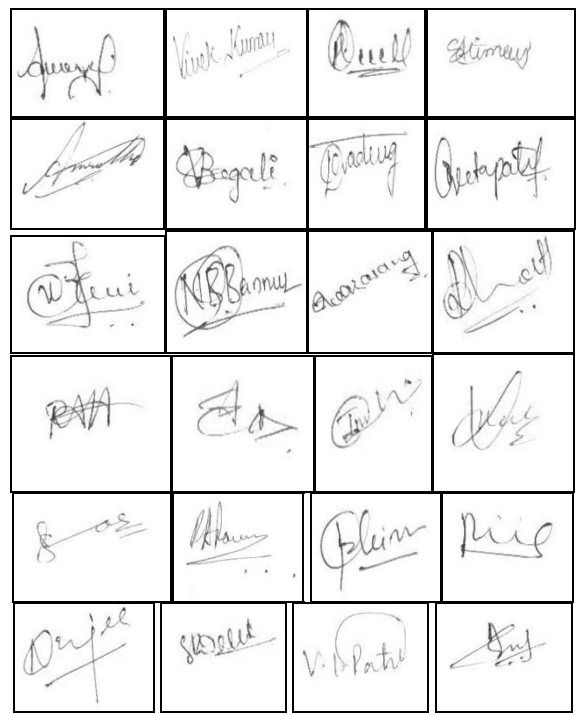

Fig.3. Sample Signature Images Database

### 4.2. Identification Performance

For each person 12 signatures for training and 4 signatures for testing are used. This makes the total of 4x52=208 signature. The identification rate is 90.6% using proposed method and 61.45 % using DWT. Fig.4 shows comparison between DWT and proposed method. From Fig. 4, we observed that signature identification rate of proposed method is superior over DWT.

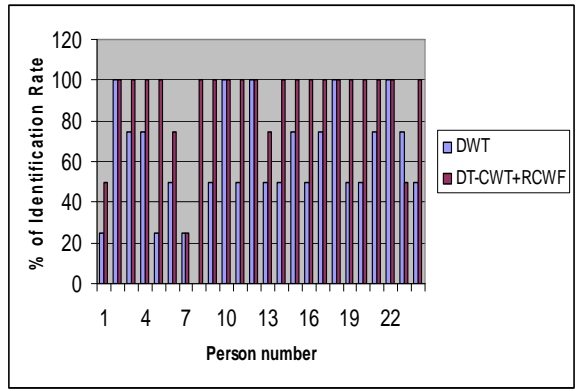

Fig.4. Comparison between DWT and RCWF

## 5. Conclusions

In this paper, we introduced new approach for identification of off-line signatures. The proposed approach uses RCWF and DT-CWT jointly for extracting details in twelve different directions and Canberra distance for comparing features. The experimental results we found that signature identification rate for proposed method is superior over DWT.

#### Acknowledgments

The authors would like to appreciate all participants who gave permission to use their handwritten signatures in this study.

**M. S. Shirdhonkar** completed his B. E. and M.E. from the Department of Computer Science and Engineering, Shivaji University, Kolhapur, India in the years 1994, 2005 respectively. From 1997-1999, he was worked as lecturer in Computer Science Department at JCE, Institute of Technology, Junner, Maharashtra, India. In 2000, he joined as a lecturer in the Department of Computer Science at B. L. D. E' s. Dr. V. P. P.G.H. College of Engineering and Technology, Bijapur, Karnataka, India, where he is presently holding position of Assistant Professor and doing PhD at S.R.T.M. University, Nanded, Maharashtra, India. His research interests include image processing, pattern recognition, and document image retrieval. He is a life member of Indian Society for Technical Education and Institute of Engineers.

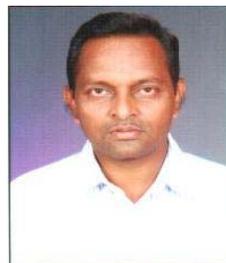

**Manesh Kokare** (S'04) was born in Pune, India, in Aug 1972. He received the Diploma in Industrial Electronics Engineering from Board of Technical Examination, Maharashtra, India, in 1990, and B.E. and M. E. Degree in Electronics from Shri Guru Gobind Singhji Institute of Engineering and Technology Nanded, Maharashtra, India, in 1993 and 1999 respectively, and Ph.D. from the Department of Electronics and Electrical Communication Engineering, Indian Institute of Technology, Kharagpur, India, in 2005. Since June1993 to Oct1995, he worked with Industry. From Oct 1995, he started his carrier in academics as a lecturer in the Department of Electronics and Telecommunication Engineering at S. G. G. S. Institute of Engineering and Technology, Nanded, where he is presently holding position of Assistant Professor. His research interests include wavelets, image processing, pattern recognition, and Content Based Image Retrieval.

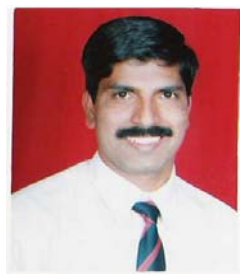